\pdfoutput=1

\documentclass[11pt]{article}

\usepackage{acl}
\usepackage{times}
\usepackage{latexsym}

\usepackage[T1]{fontenc}

\usepackage[utf8]{inputenc}

\usepackage{microtype}

\usepackage{inconsolata}

\hypersetup{
pdftitle={Language Models Can Learn Exceptions to Syntactic Rules},
pdfsubject={Language Models Can Learn Exceptions to Syntactic Rules},
pdfauthor={Cara Su-Yi Leong and Tal Linzen}
}

\title{Language Models Can Learn Exceptions to Syntactic Rules}
\usepackage{enumitem}
\usepackage[normalem]{ulem}
\usepackage{fancyhdr}
\usepackage{booktabs}
\usepackage{longtable}
\usepackage{multirow}
\usepackage{float}
\usepackage{expex}
\usepackage{todonotes}
\usepackage{amsmath,amssymb}
\usepackage{svg}
\usepackage{titlesec}
\usepackage{appendix}
\usepackage{natbib}

\usepackage{xcolor}
\definecolor{advantage}{HTML}{CC6677} 
\definecolor{estimation}{HTML}{6699CC}
\definecolor{price}{HTML}{117733}
\definecolor{duration}{HTML}{332288}
\definecolor{ooze}{HTML}{AA4499}
\definecolor{agt-pat}{HTML}{44AA99}
\definecolor{exp-th}{HTML}{DDCC77}

\newcommand{\gap}[1][0.2in]{\underline{\hspace*{#1}}}

\author{Cara Su-Yi Leong \quad Tal Linzen \\
New York University \\
  \texttt{\{caraleong,linzen\}@nyu.edu} \\}

\begin{document}
\maketitle
\begin{abstract}
Artificial neural networks can generalize productively to novel contexts. Can they also learn exceptions to those productive rules? We explore this question using the case of restrictions on English passivization (e.g., the fact that ``The vacation lasted five days'' is grammatical, but ``*Five days was lasted by the vacation'' is not). We collect human acceptability judgments for passive sentences with a range of verbs, and show that the probability distribution defined by GPT-2, a language model, matches the human judgments with high correlation. We also show that the relative acceptability of a verb in the active vs. passive voice is positively correlated with the relative frequency of its occurrence in those voices. These results provide preliminary support for the entrenchment hypothesis, according to which learners track and uses the distributional properties of their input to learn negative exceptions to rules. At the same time, this hypothesis fails to explain the magnitude of unpassivizability demonstrated by certain individual verbs, suggesting that other cues to exceptionality are available in the linguistic input.
\end{abstract}
\section{Introduction}
Many studies have demonstrated language models' ability to extend a generalization from a small set of examples to novel lexical items, structures, and contexts, even if the models do not always do so in a human-like way \citep{hupkes2020,kim-linzen-2020-cogs,lake2018,mccoy2018}. These studies show that models can substitute novel lexical items into rules where those items were previously unseen. At the same time, language models can sometimes \textit{over-generalize}, for instance by producing a literal, compositional translation of idiomatic expressions like \textit{kick the bucket} when humans would not \citep{dankers2022}. A full evaluation of language models' generalization abilities should thus not only measure whether models can generalize when humans do, but also whether models are able to \emph{constrain} their generalizations when humans do.

We address this question by building on a line of work that probes whether human-like acceptability judgments for argument structure alternations can be predicted from the probability distribution that a from language model defines over sentences. This studies have shown, for example, that the GPT-2 language model \citep{radford2019} can match human judgments about whether the dative alternation applies to a verb \citep{hawkins2020}, and that information about which syntactic frames a verb can appear in (e.g. whether a verb participates in the \textsc{spray/load} alternation) can be recovered from the verb's contextualized representations and from sentence embeddings \citep{kann-etal-2019-verb}.

In this work, we evaluate models' ability to identify exceptions using the case study of the English passive.\footnote{Data and code are available at \url{https://github.com/craaaa/exceptions}.} The passive voice is highly productive in English; most strikingly, young children exposed to novel verbs in the active voice are able to understand and produce passive constructions using those verbs \citep{pinker1987,brooks1999}. This suggests that English speakers do not in general conclude that verbs that they have never encountered in the passive voice are unacceptable in that voice. 
Yet there are limits to the productivity of the English passive; examples such as (\nextx) have been reported to be unacceptable in the passive voice:
\pex
\a The vacation lasted five days.
\a\ljudge{*} Five days was lasted by the vacation.
\xe
\begin{figure*}[ht]
    \centering
    \includegraphics[width=\textwidth]{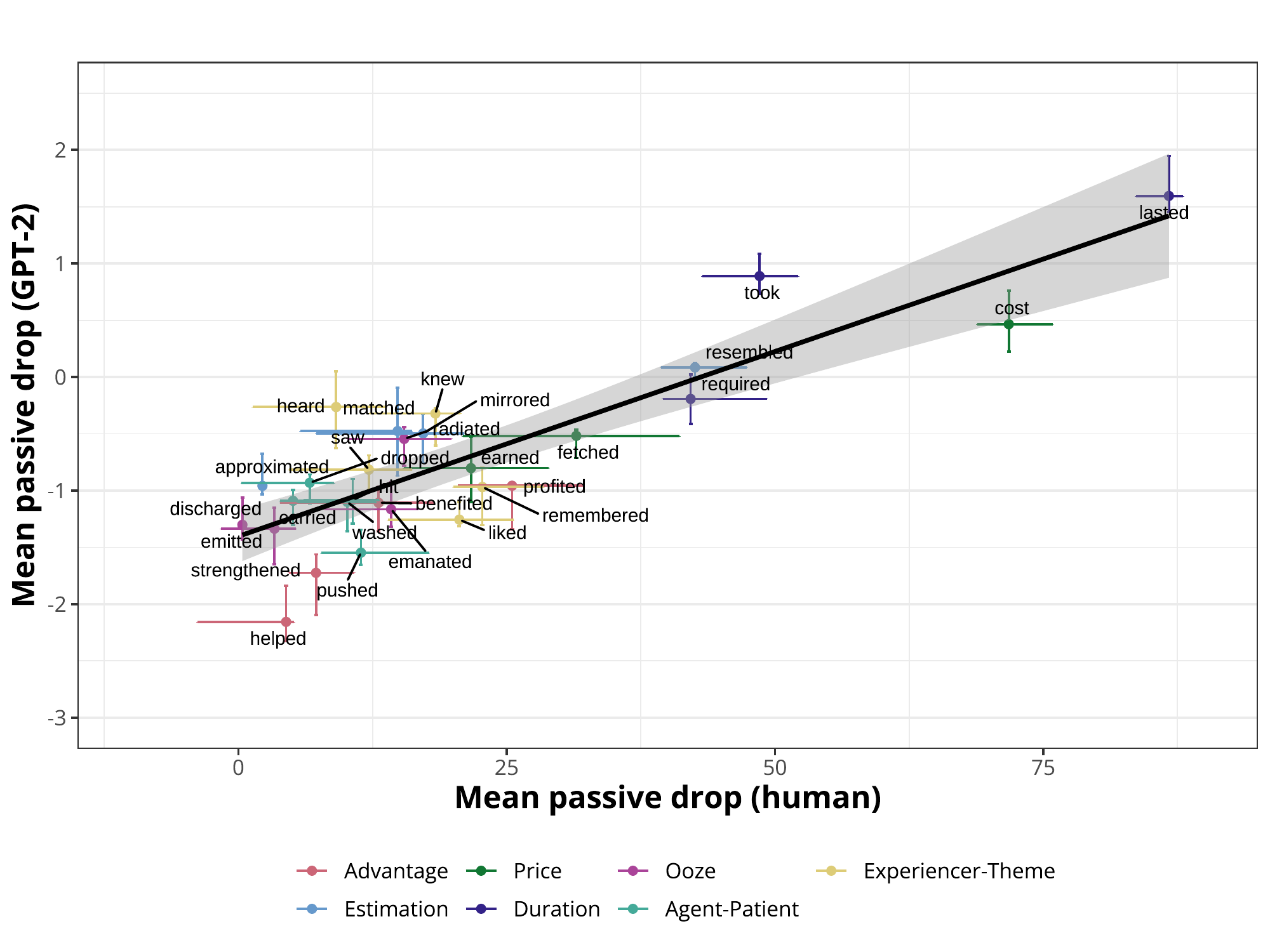}
    \caption{\textit{Passive drop in humans vs. GPT-2} --- A GPT-2 model trained on 100M words approximately predicts variable amounts of passive drop equivalent to human judgments. Horizontal and vertical error bars indicate  bootstrapped 95\% confidence intervals.}
    \label{fig:summary_scatter}
\end{figure*}

\noindent 
Sentences like (\lastx b) are unlikely to occur productively in natural speech---just like passives of infrequent verbs. Yet even though they do not receive explicit evidence that these sentences are unacceptable, rather than simply rare, English speakers nonetheless learn that they constitute exceptions, and do not judge (\lastx b) to be acceptable.

How do humans acquire such exceptions? The \textbf{entrenchment hypothesis} suggests that speakers track and use the distributional properties of their input as indirect negative evidence for the existence of an exception \citep{braine1995,regier2004,theakston2004}. For instance, if an English learner never encounters the verb \textit{last} in the passive voice, despite having seen \textit{last} used productively in the active voice, they may conclude that \textit{last} cannot occur in the the passive voice. 
Are language models---which do not have access to human feedback or syntactic supervision, and are trained solely to perform next-word prediction---attentive to the same information that humans are when determining the extent to which syntactic rules can generalize?

In this paper, we tackle these questions by comparing human acceptability judgments on sentences containing verbs that are exceptional in the passive voice, on the one hand, to the probability distribution defined by a GPT-2-like model trained on a 100-million word English corpus. We find that the language model matches human acceptability judgments on active and passive sentences to a large degree (Figure \ref{fig:summary_scatter}), suggesting that language models can constrain their syntactic generalizations in a human-like way.
Using our model's training corpus, we further show that there is a weak but positive correlation between the relative frequency of actives and passives in the input and their relative acceptability. Together, these empirical results suggest that the linguistic input contains useful information from which exceptions to syntactic generalizations can be learned.

\begin{table*}[ht]
    \centering
    \begin{tabular}{lll}
        \toprule
        Verb class & Active sentence & Passive sentence\\
        \midrule
         \textcolor{advantage}{Advantage} & Your investment \gap{} the community. & The community was \gap{} by your investment. \\
         \textcolor{price}{Price} & Your book \gap{} thirty dollars. & Thirty dollars was \gap{} by your book.\\
         \textcolor{ooze}{Ooze} & That machine \gap{} a sound. & A sound was \gap{} by that machine. \\
         \textcolor{duration}{Duration} & The journey \gap{} three days. & Three days was \gap{} by the journey.\\
         \textcolor{estimation}{Estimation} & Your drawing \gap{} her likeness. & Her likeness was \gap{} by your drawing.\\
         \bottomrule
    \end{tabular}
    \caption{\textit{Example sentence frames} --- Each verb in the verb class was substituted into frames specific to the class.}
    \label{tab:sentence_frames}
\end{table*}

\section{Restrictions on passivization}

Although the English verbal passive is highly productive, not all verbs can occur in the passive. For instance, intransitive and middle verbs resist passivization in general \citep{perlmutter1978,zaenen1993}. In this paper, we focus on passives of transitive verbs that occur with by-phrases. These long passives are clauses of the form given in (\nextx), which in most cases have an uncontroversially acceptable passive form:

\pex
\a
The ball was hit by the boy.
\xe

A small list of lexical exceptions have been described for which the passive voice is deemed ungrammatical \citep{levin1993,postal2004}. Some of these exceptions can be classed together based on the semantics of the verb or types of arguments the verb takes. For instance, verbs that take measure phrases as their object reportedly do not occur in the passive:

\pex
\a That house costs fifty thousand dollars.
\a\label{cost-pass}\ljudge{*} Fifty thousand dollars is/are cost by that house.
\trailingcitation{\citep[17-8]{hale1997}}
\xe

Even within a particular verb class, passivizability may also be an idiosyncratic characteristic of individual lexical items \citep{zwicky1987}: verbs which can be substituted for each other in any other syntactic context may differ in their ability to passivize. Thus, for instance, although in the active voice \textit{matched}, \textit{mirrored}, \textit{approximated} and \textit{resembled} can occur in the same environment, (\nextx a) is grammatical, while (\nextx b) is not.
\pex
\a Kim is matched/mirrored/approximated by the model in nearly every detail.
\a\ljudge{*} Kim is resembled by the model in nearly every detail.\trailingcitation{\citep{zwicky1987}}
\xe

\noindent We may thus expect differences in passivizability not only between verbs with different semantics and argument frames, but also among verbs with very similar meaning.

\section{Human Acceptability Judgments}
In order to test whether language models follow a human-like generalization patterns, we need to first characterize the human judgment pattern, which will serve as the target of modeling. In this section, we report on an acceptability judgment study whose goal was to verify the judgments from the syntax literature and measure any gradient differences in the degree to which different verbs can be passivized.

\subsection{Materials}
We identified five verb classes containing verbs that have been reported to be unpassivizable \citep{levin1993,postal2004,zwicky1987}:
\begin{itemize}[leftmargin=*,noitemsep]
    \item \textcolor{advantage}{Advantage} verbs: \textit{benefit}, \textit{help}, \textit{profit}, \textit{strengthen}
    \item \textcolor{price}{Price} verbs: \textit{cost}, \textit{earn}, \textit{fetch}
    \item \textcolor{ooze}{Ooze} verbs: \textit{discharge}, \textit{emanate}, \textit{emit}, \textit{radiate}
    \item \textcolor{duration}{Duration} verbs: \textit{last},  \textit{require}, \textit{take}
    \item \textcolor{estimation}{Estimation} verbs: \textit{approximate}, \textit{match}, \textit{mirror}, \textit{resemble}
\end{itemize}

\noindent Each of these class includes verbs with similar semantics that can be substituted into the same position in a sentence in the active voice. While some of these verbs can be used in other senses, we tested the specific sense that was reported in the literature by controlling the sentence frames used. Five past-tense sentence frames were constructed for each verb class (Table \ref{tab:sentence_frames}).

Each of the verbs in the class was substituted into the sentence frame, resulting in 90 total test sentence pairs. Example (\nextx) demonstrates a sentence pair generated from the sentence frame in Table \ref{tab:sentence_frames} using the verb \textit{matched}:

\pex
\a Your friend \uline{matched} my brother.
\a My brother was \uline{matched} by your friend.
\xe

As control verbs, we also selected five agent-patient and five experiencer-theme verbs;  we expect these verbs to be passivizable:
\begin{itemize}[leftmargin=*,noitemsep]
    \item \textcolor{agt-pat}{Agent-Patient}: \textit{hit}, \textit{push}, \textit{wash}, \textit{drop}, \textit{carry}
    \item \textcolor{exp-th}{Experiencer-Theme}: \textit{see}, \textit{hear}, \textit{know}, \textit{like}, \textit{remember}
\end{itemize}

\noindent Because of the varied semantics of the verbs in these groups, unique sentence pairs were created for each verb, yielding 50 control sentence pairs. An example of a sentence pair for the verb \textcolor{agt-pat}{\textit{push}} is given in (\nextx):

\pex
\a A boy \uline{pushed} the cup.
\a The cup was \uline{pushed} by a boy.
\xe

Each participant only saw either the active or the passive of a sentence pair.
The 140 sentence pairs (90 test + 50 control) were divided into two buckets of 70 sentence pairs each such that each bucket contained two or three sentence frames per verb. Each bucket was then further divided into groups of 70 sentences such that the active and passive forms of a sentence pair were in different groups. Each group of sentences contained one quarter of the test and control stimuli (70 sentences).

Presentation order was counterbalanced by making four ordered lists for each group. Each group was organized into two lists such that an item that appeared in the first half of of one list appeared in the second half of the other list. The order of items was pseudorandomized within those lists to ensure that not more than two active or passive sentences and no two sentences within the same verb class were seen in succession. These lists were then reversed, so that a total of four ordered sentence lists were made per sentence group.

Additionally, every experimental trial alternated with a filler sentence. Filler sentences were also used as attention checks. We used 24 grammatical and 46 ungrammatical filler sentences: since the passives of control sentences were expected to be acceptable, the greater number of ungrammatical fillers was intended to balance the experimental stimuli. The full set of materials is available in Appendix \ref{sec:stimuli}.

\begin{figure*}[ht]
    \centering
    \includegraphics[width=\textwidth]{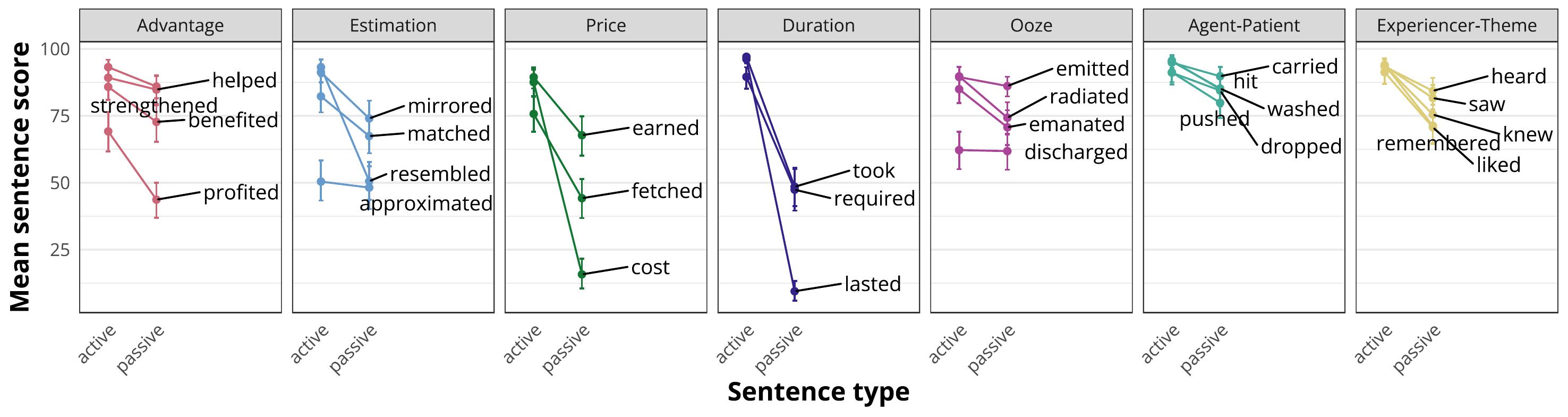}
    \caption{\textit{Passive drop in human acceptability judgments of active and passive sentences by verb} --- The steeper the downward gradient between active and passive conditions, the larger the passive drop. Error bars indicate bootstrapped 95\% confidence intervals.}
    \label{fig:all_verbs_duckbill}
\end{figure*}

\subsection{Participants}
We recruited 84 participants who had IP addresses located in the US and self-reported as native English speakers via the crowdsourcing platform Prolific. Each participant rated 140 sentences (70 test + 70 filler) and was paid US\$3.50. The experiment took approximately 12 minutes to complete.

Participants were asked to rate how acceptable each sentence sounded based on their gut reaction. They were told that there were no right or wrong answers. Participants rated sentences by moving a slider from ``Completely unacceptable" to ``Completely acceptable", which corresponded to an integer score (invisible to them) between 0 and 100. They were not able to rate a sentence with a score of 50. Two practice sentences (one ungrammatical, one grammatical) were used to familiarize participants with the paradigm.

Participants were excluded from the results if they answered more than 15 filler questions unexpectedly, either by giving ungrammatical sentences scores above 50 or giving grammatical sentences scores below 50. We excluded 10 participants from analysis for this reason.

\subsection{Results} 
We calculate the \textbf{passive drop} of an item as the difference in mean acceptability ratings between its active and passive version. The results are reported in Figure~\ref{fig:all_verbs_duckbill}; a steeper downward gradient corresponds with a larger passive drop. 
Since corresponding active and passive sentences contain the same lexical items except for the auxiliary \textit{was} and \textit{by}, which are common across all sentences, directly comparing active and passive sentences isolates the effect of the passivization from lexical effects that might increase the acceptability of sentences with more common verbs like \textit{helped} over low-frequency verbs like \textit{profited}.

Across all verb classes, there was a significant difference between scores given to active and passive sentences. This difference may be accounted for by pragmatic factors: although the passive construction is more pragmatically marked than the active \citep{comrie1988}, each sentence in the acceptability judgment task was presented to participants without establishing a relevant context. This setting might have caused participants to rate passive sentences as worse than their active counterparts.

Although the passive drop was positive for all verbs, its \textbf{magnitude} differed across verb classes. The \textcolor{duration}{duration} class showed the largest mean passive drop (59.4 points), and the \textcolor{ooze}{ooze} class showed the lowest mean passive drop (8.0 points) among the test verb classes. 

We fit a linear mixed-effects model to predict \textsc{sentence score} using the \textcolor{agt-pat}{agent-patient} verb class as the baseline. We used \textsc{sentence type} and \textsc{verb class} as well as their interaction as fixed effects and \textsc{frame}, \textsc{participant} and \textsc{verb} as random intercepts.
We found a significant difference between \textcolor{agt-pat}{agent-patient} verbs and three other verb classes: \textcolor{estimation}{estimation} verbs ($p=$ 5.74e-06), \textcolor{price}{price} verbs ($p<$ 2e-16), and \textcolor{duration}{duration} verbs ($p<$ 2e-16). On the other hand, there was no significant difference in the sentence scores obtained from \textcolor{agt-pat}{agent-patient} verbs and \textcolor{ooze}{ooze} verbs, \textcolor{advantage}{advantage} verbs, or \textcolor{exp-th}{experiencer-theme} verbs as a class.

Within each verb class, some verbs were more passivizable than others. For example, \textcolor{duration}{\textit{last}} was significantly less passivizable than \textcolor{duration}{\textit{took}} and \textcolor{duration}{\textit{required}}, and \textcolor{price}{\textit{cost}} was less passivizable than \textcolor{price}{\textit{fetched}}. Similarly, while \textcolor{estimation}{\textit{resembled}} had a high passive drop, the remaining verbs in the \textcolor{estimation}{estimation} class showed relatively low passive drops. 
These results validate the claim that some verbs may be more passivizable than others despite sharing similar paradigmatic relationships \citep{zwicky1987}.

In summary, the human acceptability judgment experiment demonstrated that some verbs in the verb classes being tested are degraded in the passive voice, and that unacceptability was gradient between verbs. For a model to adequately approximate such behaviour, it must exhibit the following characteristics:

\vspace{-.5em}
\begin{itemize}[noitemsep]
    \item \textbf{Exceptionality}: some verbs (e.g. \textcolor{duration}{duration} verbs) exhibit passive drops that are significantly different from the baseline passive drop expected of the canonically passivizable \textcolor{agt-pat}{agent-patient} verbs.
    \item \textbf{Gradience}: (un)acceptability is gradient, with some verbs on average exhibiting higher passive drop than others.
\end{itemize}

\section{Comparison with Language Models}
With the quantitative human acceptability judgment data in hand, we now turn to evaluating language models.
If distributional data is sufficient to learn the extent to which verbs are unacceptable in the passive, we expected GPT-2 to be able to match human judgments on both passivizable verbs and unpassivizable verbs. We also expect GPT-2 to be able to match the relative gradience of passive drop that humans display.

\subsection{Method}
We evaluated GPT-2 \citep{radford2019}, a Transformer \citep{vaswani2017} language model. 
We tested four different pre-trained GPT-2 models, which differed in their number of parameters and number of layers, but were trained on the same data. Each model was trained on Open-AI's WebText corpus, which contains 40GB of data --- approximately 8B words, assuming each word contains an average of 5 bytes/chars. Pre-trained GPT-2 models have performed well on targeted syntactic evaluations requiring knowledge of argument structure, such as differentiating between verbs that participate in the causative alternation and those that do not \citep{warstadt2020}.

The GPT-2 models available for download are trained on a much larger corpus than is realistic for any human to be exposed to \cite{linzen2020}. English-speaking children are exposed to 2--7M words per year \citep{gilkerson2017}, or 26M--91M words by the age of 13. Rounding to the nearest order of magnitude, we trained a GPT-2 model on a 100M word subset of the OpenWebText corpus \citep{gokaslan2019}, an open-source reproduction of the Web Text corpus; this simulates more closely the amount of linguistic input a human may receive (though not its genre). We trained five iterations of this model, which we call \textbf{GPT2-100M}, using different random seeds and report averages of the results obtained from these five models.

We adapted the  targeted syntactic evaluation paradigm \citep{linzen2016,lau2017,warstadt-etal-2019-neural} to compare the language models to humans. This paradigm involves obtaining model ``judgments'' for minimal pairs of sentences. For each sentence, a score is obtained by summing the log-probabilities assigned to each token in the sentence, which gives the probability the model assigns to that sentence. We conclude that a model's distribution is consistent with human judgments if it assigns a higher probability to the acceptable sentence than to the corresponding unacceptable one. Unlike some prior work, we collected numeric scores instead of binary acceptability judgments: we calculated a gradient passive drop of each sentence pair by subtracting the score of the active sentence from the score of its passive counterpart.

Since we compared active sentences to long passives, which contain by-phrases, every passive sentence contained two more words than its active counterpart. A sentence with more tokens will on balance be less probable than a sentence with fewer tokens; we thus normalized each sentence score by dividing it by the number of tokens in the sentence \cite{lau2017}. Doing so accounts for the effect of sentence length on the sentence score, and also allows us to compare sentences where words are split into separate tokens during GPT-2's tokenization process, e.g. approximated $\rightarrow$ approx + imated.

\subsection{Results}
The four pre-trained models as well as the five GPT2-100M models showed positive correlations between mean human passive drop and mean model passive drop, reported in Table \ref{tab:models}. For pre-trained GPT-2 models, we calculate mean model passive drop for each verb by averaging over the passive drop of all five sentence frames. For GPT2-100M, we calculate the average passive drop of each verb over all sentence frames across the five versions of the model (trained with different random seeds); we report these results as \textbf{GPT2-100M-avg}. 

\begin{figure}[t]
    \centering
    \includegraphics[width=\columnwidth]{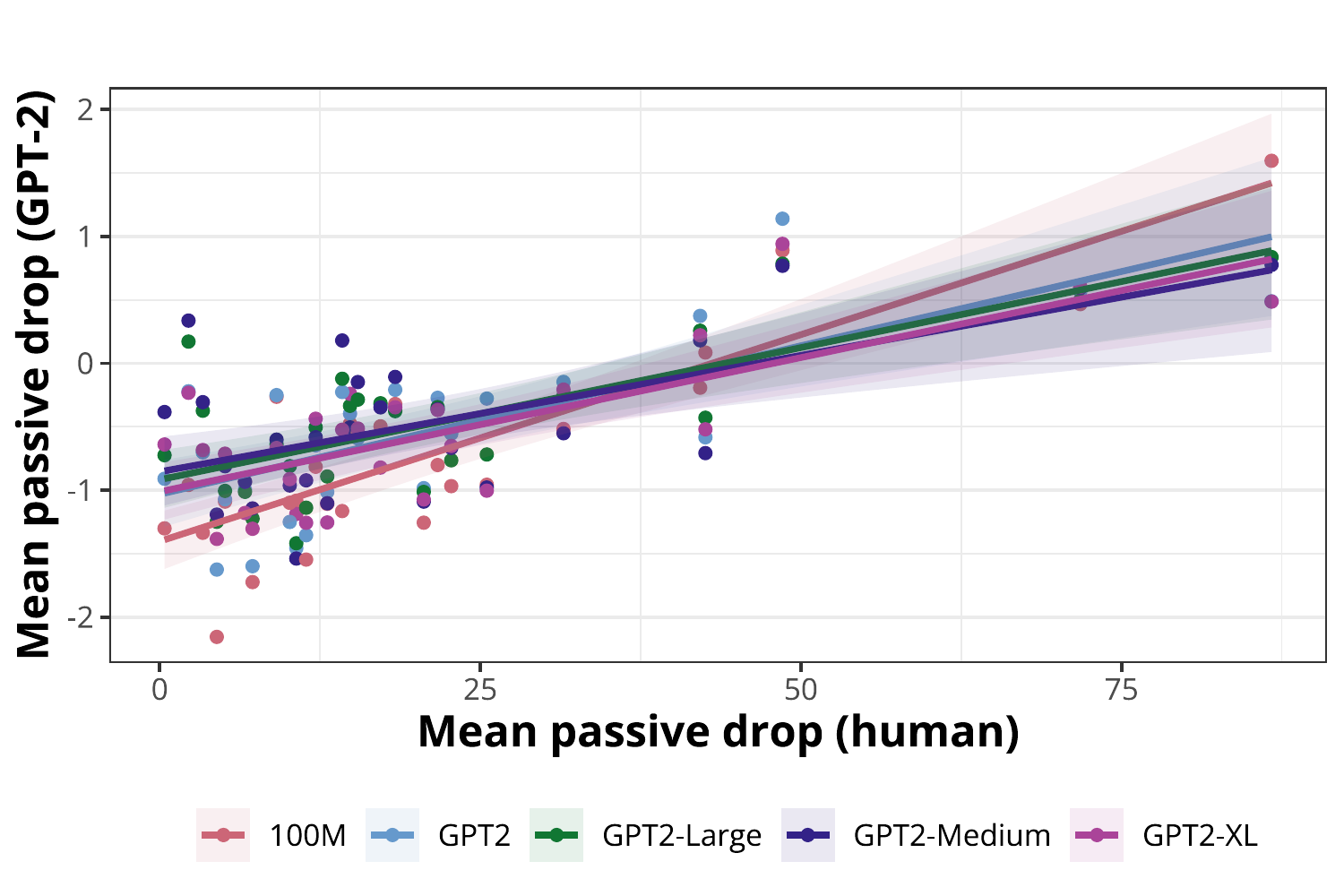}
    \caption{\textit{Passive drop of different-sized GPT-2 models compared to human judgments} --- Each point in represents a single verb. Models differed in number of parameters and/or training data, but showed qualitatively similar passive drops.}
\label{fig:model_size}
\end{figure}

The results were qualitative similar for all models (Figure \ref{fig:model_size}); in what follows, we focus on GPT2-100M-avg, whose behaviour showed the strongest correlation with human judgments. These models are also trained on the most cognitively realistic corpus.

\begin{table}[t]
\centering
\begin{tabular}{lll}
    \toprule
    Model & \# parameters & $r_s$\\
    \midrule
     GPT2-100M-avg & 124M & \textbf{0.709} \\
     GPT2 &  124M & 0.659\\
     GPT2-med & 345M & 0.385\\
     GPT2-large & 774M & 0.549\\
     GPT2-xl & 1558M & 0.559\\
     \bottomrule
\end{tabular}
\caption{\textit{GPT2 model parameters and correlation coefficients} --- in all five models, a correlation was found between human passive drop and the model's passive drop, but it was stronger for smaller models, and strongest for the models trained on only 100M words.}
\label{tab:models}
\end{table}

Figure \ref{fig:summary_scatter} plots GPT2-100M-avg's passive drop against the passive drop observed in the human experiment. A strong correlation was found between the passive drop in the models' sentence scores and human passive drop ($r_s$ = 0.709), suggesting that predictions learned from linguistic input match human \textbf{gradient} judgments on passivization relatively well.

\begin{figure*}[ht]
    \centering
    \includegraphics[width=\textwidth]{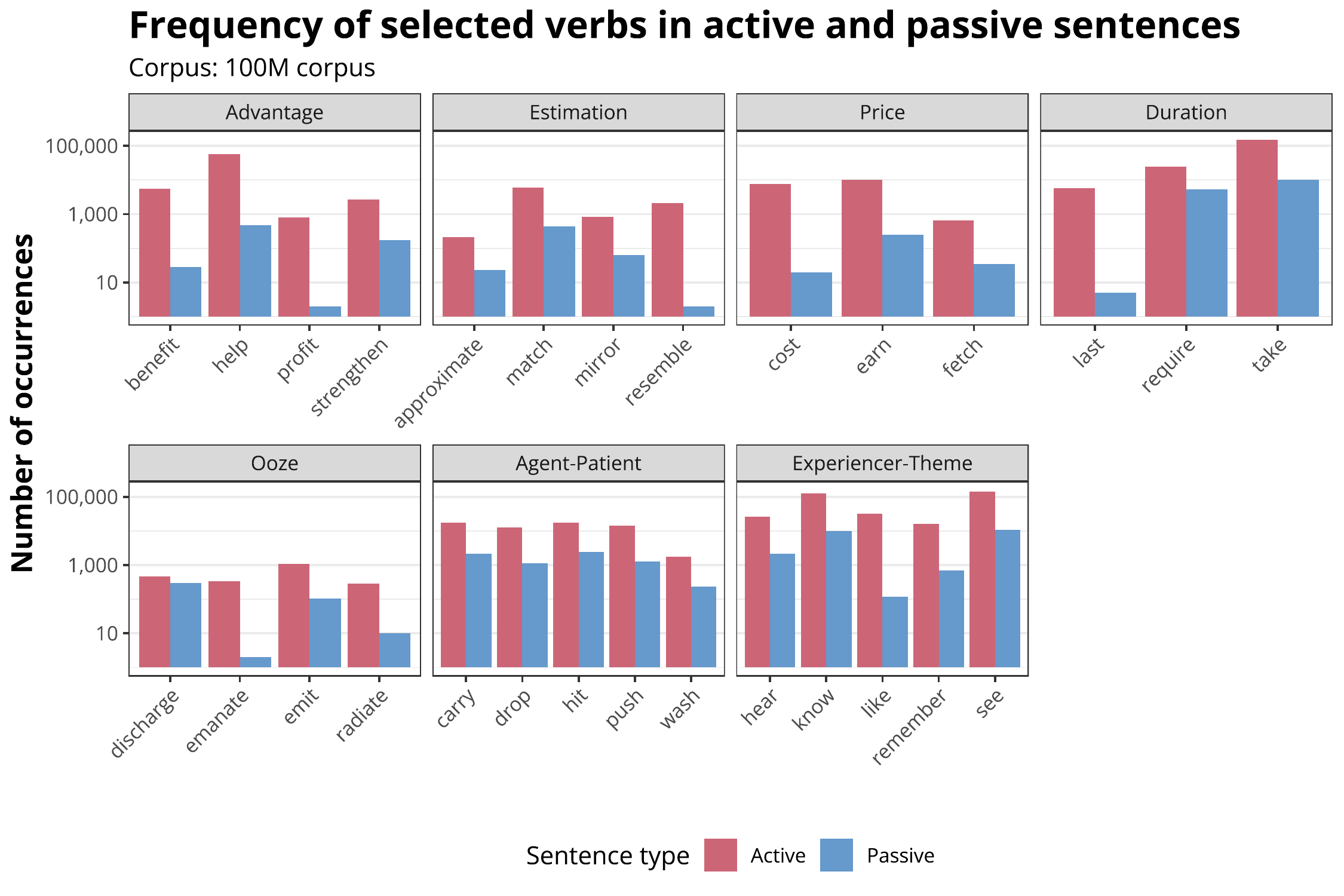}
    \caption{\textit{Occurrence of active transitive and passive sentences using test verbs in the training corpus} --- sentences whose verb had a passive dependent (\texttt{csubjpass}, \texttt{nsubjpass}, or \texttt{auxpass}) were tagged as passive, while all other instances of the verb were tagged as active.}
    \label{fig:training_data}
\end{figure*}

GPT2-100M-avg also matched humans' judgments of \textbf{exceptionality} within verb classes: among verbs with similar meanings, both humans and the model identified the same verbs as being less passivizable.
In verbs for which humans demonstrated low passive drop, such as \textcolor{advantage}{\textit{strengthened}} and \textcolor{ooze}{\textit{discharged}}, close to no passive drop was observed in the model's predictions.
GPT2-100M-avg also predicted high passive drops for verbs like \textcolor{duration}{\textit{lasted}}, \textcolor{estimation}{\textit{resembled}} and \textcolor{price}{\textit{cost}}, aligning with  human judgments that these verbs are unique in their verb class.

\section{Does Frequency Explain Passivizability Judgments?}

Having established that a language model can successfully model humans' gradient passivizability judgments, we now examine the extent to which GPT2-100M's passivization judgments  correlate with the distributional properties of its training data. Specifically, we explore the utility of the entrenchment hypothesis in explaining GPT2-100M's gradient judgments of passivization. Recall that this hypothesis argues that learners conclude that a verb cannot appear in a particular context if it appears in many other contexts but systematically fails to appear in the context in question.

Here, we consider a weaker version of the entrenchment hypothesis, which does not presuppose that exceptions \textit{never} occur in the learner's input. Instead, we hypothesize that the less frequently a verb is used in the passive voice relative to the active voice, the less acceptable passive constructions using that verb will be.

\begin{figure*}[ht]
    \centering
    \includegraphics[width=\textwidth]{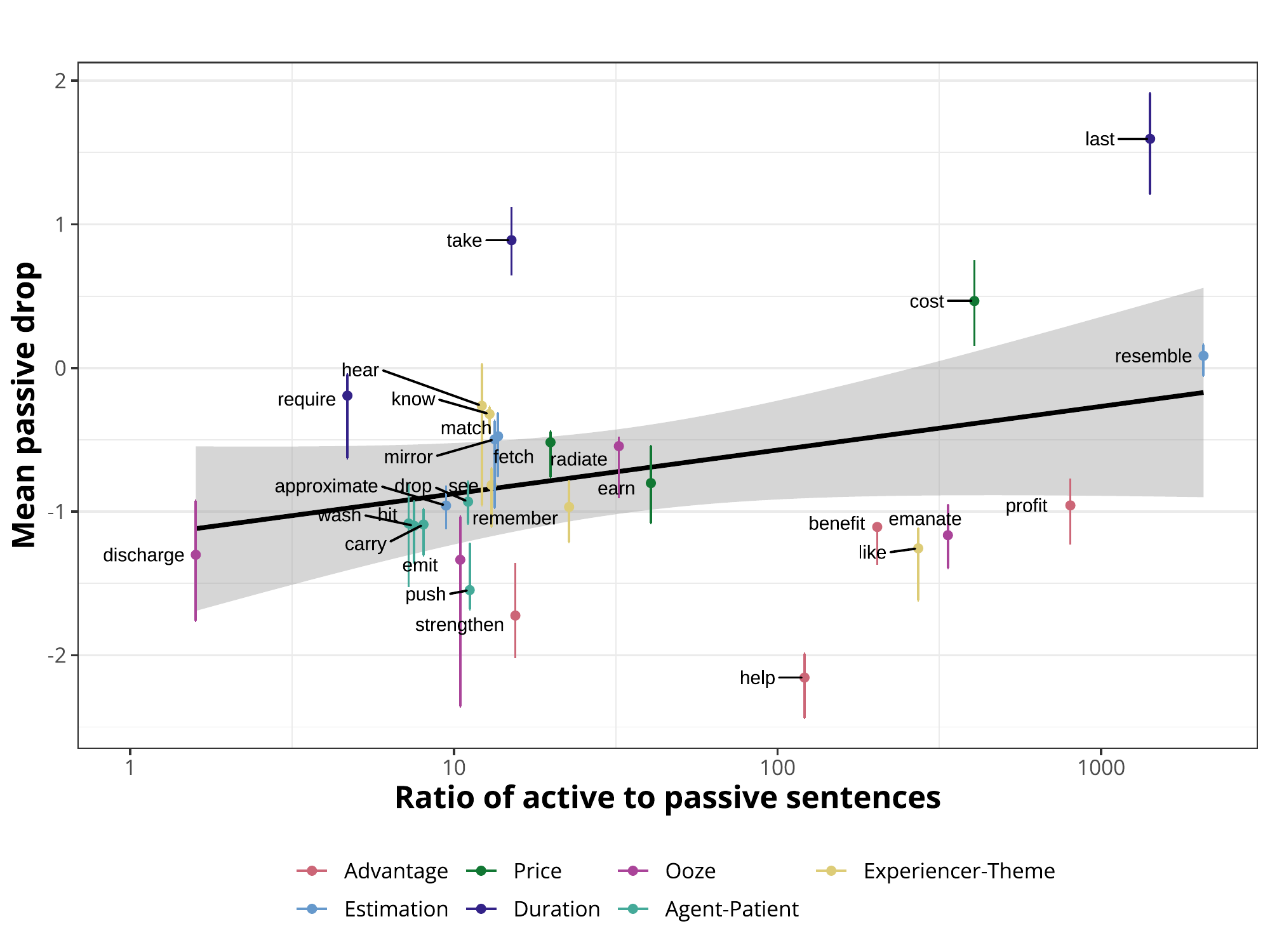}
    \caption{\textit{GPT2-100M's passive drop against the ratio of active to passive sentences in its training corpus}. Error bars indicate bootstrapped 95\% confidence intervals across sentence frames.}
    \label{fig:frequency_correlation}
\end{figure*}

\subsection{Method}
We conducted a corpus study on the data that GPT2-100M was trained on. We processed each document in the corpus using the spaCy Transformer-based lemmatizer, POS tokenizer and dependency parser \citep{honnibal2020} and extracted all sentences that contained a verbal lemma corresponding to the test and control verbs. Sentences that contained the verbs in question and had a dependency edge to a passive auxilliary (\texttt{auxpass}), a passive nominal subject (\texttt{nsubjpass}) or a passive clausal subject (\texttt{csubjpass}) were classified as passive sentences, while all other sentences containing the verb were classified as active sentences. We hand-checked a 1000 sentence subset of the training data to verify the accuracy of the tagging process. No sentences were incorrectly tagged in the manually verified subset, although the corpus did contain instances of typos such as (\nextx) (tagged as passive):

\ex
It was fun while it was lasted.
\xe

\subsection{Results}

Figure \ref{fig:training_data} shows the number of active and passive sentences in GPT2-100M's training corpus. 

Not all verbs appear in the same ratios in the active and passive voice. Agent-patient verbs consistently appeared in approximately 10 times as many active sentences as passive sentences, matching estimates from previous corpus studies \citep{roland2007}. On the other hand, test verbs appeared in varying amounts in the active and passive. For instance, \textit{last} appeared 5666 times in the active and four times in the passive in the 100M word corpus, while \textit{cost} appeared 7706 times in the active and 19 times in the passive.
This result suggests that the test verbs differ from canonically passivizable control verbs in their distribution.

Figure \ref{fig:frequency_correlation} graphs the correlation between the ratio of active to passive sentences for a given verb, on the one hand, and that verb's mean passive drop on the other hand. We find a weak but positive correlation between the two variables ($r_s$= 0.212). 

Two key outliers that are not well accounted for by this measure of relative frequency are \textcolor{duration}{\textit{last}} and \textcolor{price}{\textit{cost}}. In both humans and model judgments, these verbs demonstrated high passive drops; yet, they are similar in relative frequency of active and passive to verbs like \textcolor{ooze}{\textit{emanate}}, \textcolor{advantage}{\textit{profit}} and \textcolor{estimation}{\textit{resemble}}, whose passive drops are lower. While frequency seems to predict some amount of unpassivizability, then, it cannot account for the full magnitude of the passive drop displayed by these particular verbs.

Furthermore, entire verb classes are systematically over- or under-predicted in Figure \ref{fig:frequency_correlation}. The \textcolor{duration}{duration} verb class on the whole has a high passive drop relative to its frequency in the corpus, while frequency over-predicts the passive drop expected for the \textcolor{advantage}{advantage} verb class. We thus conclude that while the relative frequency of active and passive voice sentences positively correlates with passive drop, other factors are likely to also be relevant on a verb-class level.

Although \textcolor{duration}{\textit{take}} appears to be an outlier in Figure~\ref{fig:frequency_correlation}, with an active to passive ratio similar to that of the \textcolor{agt-pat}{agent-patient} and \textcolor{exp-th}{experiencer-theme} control verbs, the measure of frequency we used does not take into account the fact that \textit{take} has multiple senses. If a different sense than the one being tested is heavily represented by passive sentences, the number of passives counted may be overestimated. For example, although we only test the duration sense of \textit{take}, as given in (\nextx a), the sense used in (\nextx b) may be more prevalent in the corpus: 

\pex
\a\ljudge{*} Two days was \uline{taken} by the meeting. 
\a The photo was \uline{taken} by the boy.
\xe

\noindent These differences in verb sense are not accounted for in the current corpus study; future work should make use of word sense disambiguation to conduct more targeted corpus analyses. Additionally, the issue of differentiating verb senses in polysemous verbs is one that both human and machine learners face, raising the question of the extent to which learners differentiate between verb senses that are more or less difficult to passivize.

Overall, while the relative frequency of a verb's occurrence in the active and passive does positively correlate with its unpassivizability, it does not account for crucial verb-level differences in the magnitude of passive drop demonstrated by GPT2-100M-avg.
\section{Discussion}
The goal of this study was to explore whether a language model can identify exceptions to a productive syntactic rule in a human-like way. We compared human acceptability judgments to sentence scores produced by a GPT-2 model trained on the amount of linguistic input that a human can plausibly be exposed to, and found that the model displayed human-like exceptionality and gradience in its judgments of passive sentences. The results of our study suggest that language models are able to refrain from over-generalizing to exceptions. Our results suggest that future empirical inroads may be made towards understanding the mechanisms and input required to overcome the projection problem \citep{baker1979}, i.e. the problem of acquiring arbitrary negative exceptions, using language models as experimental subjects.

We took a first step in this direction by showing a positive correlation between the relative frequency of active and passive sentences containing a given verb and the difference between that verb's acceptability in the active and passive voice (i.e. its passive drop) in GPT-2. Although our results lend some credence to the entrenchment hypothesis, they suggest that additional factors must be recruited to explain the full magnitude of exceptionality displayed by highly unpassivizable verbs such as \textit{last} and \textit{cost}.

Moreover, although we demonstrated that the relative frequency of a verb's occurrence in the active and passive is correlated with its passive drop, a causal relationship between the two cannot be established from our data. A single underlying factor, such as verbal semantics, may affect both the frequency of a verb in the passive in relation to the passive \textit{and} its acceptability in the passive construction. 

Future research should test the causal impact of a verb's absolute and relative frequency in the training corpus on its predicted passivizability. Following \citet{wei-etal-2021-frequency}, we plan to create an altered training dataset where we match the frequency of active and passive sentences containing passivizable verbs like \textit{drop} to the absolute frequency of sentences containing highly unpassivizable verbs, such as \textit{last}. Comparing models trained on this dataset against GPT2-100M will allow us to move beyond a correlational analysis and explore whether altering the frequency of a verb in the active and passive voice in a model's training data has a causal effect on the model's predictions of that verb's passivizability. 

\section{Conclusion}
In this paper, we explored whether a language model trained on a human-scale amount of linguistic input is able to learn lexical exceptions to a productive syntactic generalization in English. We showed that it was able to match humans' reported judgments on unpassivizable verbs like \textit{last}, showing both the ability to identify exceptions as well as to identify the magnitude of an exception. We also demonstrated a weak correlation between the degree to which a model prefers active over passive sentences using a given verb, on the one hand, and the ratio between the frequencies with which sentences containing that verb occur in the active and passive voice, on the other hand. Together, these results suggest that distributional information plays a role in learning exceptions to syntactic rules.

\section*{Acknowledgments}
We would like to thank Alec Marantz and Gary Thoms for valuable comments and suggestions related to this paper. This material is based upon work supported by the National Science Foundation (NSF) under Grant No. BCS-2114505. This work was supported in part through the NYU IT High Performance Computing resources, services, and staff expertise.

\bibliography{passives,anthology}

\begin{thebibliography}{32}
\expandafter\ifx\csname natexlab\endcsname\relax\def\natexlab#1{#1}\fi

\bibitem[{Baker(1979)}]{baker1979}
C.~L. Baker. 1979.
\newblock \href {http://arxiv.org/abs/4178133} {Syntactic {{Theory}} and the {{Projection Problem}}}.
\newblock \emph{Linguistic Inquiry}, 10(4):533--581.

\bibitem[{Braine and Brooks(1995)}]{braine1995}
Martin D~S Braine and Patricia~J Brooks. 1995.
\newblock Verb {{Argument Structure}} and the {{Problem}} of {{Avoiding}} an {{Overgeneral Grammar}}.
\newblock In Michael Tomasello and William~Edward Merriman, editors, \emph{Beyond Names for Things: Young Children's Acquisition of Verbs}. {L. Erlbaum}, {Hillsdale, N.J}.

\bibitem[{Brooks and Tomasello(1999)}]{brooks1999}
Patricia~J. Brooks and Michael Tomasello. 1999.
\newblock \href {https://doi.org/10.1037/0012-1649.35.1.29} {Young children learn to produce passives with nonce verbs.}
\newblock \emph{Developmental Psychology}, 35(1):29.

\bibitem[{Comrie(1988)}]{comrie1988}
Bernard Comrie. 1988.
\newblock \href {https://doi.org/10.1075/tsl.16.04com} {Passive and voice}.
\newblock In Masayoshi Shibatani, editor, \emph{Passive and {{Voice}}}, Typological {{Studies}} in {{Language}}, pages 9--24. {John Benjamins Publishing Company}.

\bibitem[{Dankers et~al.(2022)Dankers, Lucas, and Titov}]{dankers2022}
Verna Dankers, Christopher Lucas, and Ivan Titov. 2022.
\newblock \href {https://doi.org/10.18653/v1/2022.acl-long.252} {Can {{Transformer}} be {{Too Compositional}}? {{Analysing Idiom Processing}} in {{Neural Machine Translation}}}.
\newblock In \emph{Proceedings of the 60th {{Annual Meeting}} of the {{Association}} for {{Computational Linguistics}} ({{Volume}} 1: {{Long Papers}})}, pages 3608--3626, {Dublin, Ireland}. {Association for Computational Linguistics}.

\bibitem[{Gilkerson et~al.(2017)Gilkerson, Richards, Warren, Montgomery, Greenwood, Kimbrough, Hansen, and Paul}]{gilkerson2017}
Jill Gilkerson, Jeffrey~A. Richards, Steven~F. Warren, Judith~K. Montgomery, Charles~R. Greenwood, Oller~D. Kimbrough, John H.~L. Hansen, and Terrance~D. Paul. 2017.
\newblock \href {https://doi.org/10.1044/2016_AJSLP-15-0169} {Mapping the {{Early Language Environment Using All-Day Recordings}} and {{Automated Analysis}}}.
\newblock \emph{American Journal of Speech-Language Pathology}, 26(2):248--265.

\bibitem[{Gokaslan and Cohen(2019)}]{gokaslan2019}
Aaron Gokaslan and Vanya Cohen. 2019.
\newblock {{OpenWebText}} corpus.

\bibitem[{Hale and Keyser(1997)}]{hale1997}
Ken Hale and Samuel~Jay Keyser. 1997.
\newblock Adjectives, other stative predicates and the roots of stativity.

\bibitem[{Hawkins et~al.(2020)Hawkins, Yamakoshi, Griffiths, and Goldberg}]{hawkins2020}
Robert Hawkins, Takateru Yamakoshi, Thomas Griffiths, and Adele Goldberg. 2020.
\newblock \href {https://doi.org/10.18653/v1/2020.emnlp-main.376} {Investigating representations of verb bias in neural language models}.
\newblock In \emph{Proceedings of the 2020 {{Conference}} on {{Empirical Methods}} in {{Natural Language Processing}} ({{EMNLP}})}, pages 4653--4663, {Online}. {Association for Computational Linguistics}.

\bibitem[{Honnibal et~al.(2020)Honnibal, Montani, Van~Landeghem, and Boyd}]{honnibal2020}
Matthew Honnibal, Ines Montani, Sofie Van~Landeghem, and Adriane Boyd. 2020.
\newblock {{spaCy}}: {{Industrial-strength}} natural language processing in python.

\bibitem[{Hupkes et~al.(2020)Hupkes, Dankers, Mul, and Bruni}]{hupkes2020}
Dieuwke Hupkes, Verna Dankers, Mathijs Mul, and Elia Bruni. 2020.
\newblock \href {https://doi.org/10.24963/ijcai.2020/708} {Compositionality {{Decomposed}}: {{How}} do {{Neural Networks Generalise}}? ({{Extended Abstract}})}.
\newblock In \emph{Proceedings of the {{Twenty-Ninth International Joint Conference}} on {{Artificial Intelligence}}}, pages 5065--5069, {Yokohama, Japan}. {International Joint Conferences on Artificial Intelligence Organization}.

\bibitem[{Kann et~al.(2019)Kann, Warstadt, Williams, and Bowman}]{kann-etal-2019-verb}
Katharina Kann, Alex Warstadt, Adina Williams, and Samuel~R. Bowman. 2019.
\newblock \href {https://doi.org/10.7275/q5js-4y86} {Verb argument structure alternations in word and sentence embeddings}.
\newblock In \emph{Proceedings of the Society for Computation in Linguistics ({SC}i{L}) 2019}, pages 287--297.

\bibitem[{Kim and Linzen(2020)}]{kim-linzen-2020-cogs}
Najoung Kim and Tal Linzen. 2020.
\newblock \href {https://doi.org/10.18653/v1/2020.emnlp-main.731} {{COGS}: A compositional generalization challenge based on semantic interpretation}.
\newblock In \emph{Proceedings of the 2020 Conference on Empirical Methods in Natural Language Processing (EMNLP)}, pages 9087--9105, Online. Association for Computational Linguistics.

\bibitem[{Lake and Baroni(2018)}]{lake2018}
Brenden~M. Lake and Marco Baroni. 2018.
\newblock Generalization without {{Systematicity}}: {{On}} the {{Compositional Skills}} of {{Sequence-to-Sequence Recurrent Networks}}.
\newblock In \emph{Proceedings of the 35th {{International Conference}} on {{Machine Learning}}, {{ICML}} 2018, {{Stockholmsm\"assan}}, {{Stockholm}}, {{Sweden}}, {{July}} 10-15, 2018}, volume~80 of \emph{Proceedings of {{Machine Learning Research}}}, pages 2879--2888. {PMLR}.

\bibitem[{Lau et~al.(2017)Lau, Clark, and Lappin}]{lau2017}
Jey~Han Lau, Alexander Clark, and Shalom Lappin. 2017.
\newblock \href {https://doi.org/10.1111/cogs.12414} {Grammaticality, {{Acceptability}}, and {{Probability}}: {{A Probabilistic View}} of {{Linguistic Knowledge}}}.
\newblock \emph{Cognitive Science}, 41(5):1202--1241.

\bibitem[{Levin(1993)}]{levin1993}
Beth Levin. 1993.
\newblock \emph{English Verb Classes and Alternations: A Preliminary Investigation}.
\newblock {University of Chicago Press}, {Chicago}.

\bibitem[{Linzen(2020)}]{linzen2020}
Tal Linzen. 2020.
\newblock \href {https://doi.org/10.18653/v1/2020.acl-main.465} {How {{Can We Accelerate Progress Towards Human-like Linguistic Generalization}}?}
\newblock In \emph{Proceedings of the 58th {{Annual Meeting}} of the {{Association}} for {{Computational Linguistics}}}, pages 5210--5217, {Online}. {Association for Computational Linguistics}.

\bibitem[{Linzen et~al.(2016)Linzen, Dupoux, and Goldberg}]{linzen2016}
Tal Linzen, Emmanuel Dupoux, and Yoav Goldberg. 2016.
\newblock \href {https://doi.org/10.1162/tacl_a_00115} {Assessing the {{Ability}} of {{LSTMs}} to {{Learn Syntax-Sensitive Dependencies}}}.
\newblock \emph{Transactions of the Association for Computational Linguistics}, 4:521--535.

\bibitem[{McCoy et~al.(2018)McCoy, Frank, and Linzen}]{mccoy2018}
R.~Thomas McCoy, Robert Frank, and Tal Linzen. 2018.
\newblock {Revisiting the poverty of the stimulus: hierarchical generalization without a hierarchical bias in recurrent neural networks}.
\newblock In \emph{{Proceedings of the 40th Annual Meeting of the Cognitive Science Society, CogSci 2018}}, {Proceedings of the 40th Annual Meeting of the Cognitive Science Society, CogSci 2018}, pages 2096--2101. {The Cognitive Science Society}.

\bibitem[{Perlmutter(1978)}]{perlmutter1978}
David~M. Perlmutter. 1978.
\newblock \href {https://doi.org/10.3765/bls.v4i0.2198} {Impersonal {{Passives}} and the {{Unaccusative Hypothesis}}}.
\newblock \emph{Annual Meeting of the Berkeley Linguistics Society}, 4(0):157--190.

\bibitem[{Pinker et~al.(1987)Pinker, Lebeaux, and Frost}]{pinker1987}
Steven Pinker, David~S. Lebeaux, and Loren~Ann Frost. 1987.
\newblock \href {https://doi.org/10.1016/S0010-0277(87)80001-X} {Productivity and constraints in the acquisition of the passive}.
\newblock \emph{Cognition}, 26(3):195--267.

\bibitem[{Postal(2004)}]{postal2004}
Paul~Martin Postal. 2004.
\newblock \emph{Skeptical Linguistic Essays}.
\newblock {Oxford University Press}, {Oxford ; New York}.

\bibitem[{Radford et~al.(2019)Radford, Wu, Child, Luan, Amodei, and Sutskever}]{radford2019}
Alec Radford, Jeffrey Wu, Rewon Child, David Luan, Dario Amodei, and Ilya Sutskever. 2019.
\newblock Language {{Models}} are {{Unsupervised Multitask Learners}}.
\newblock Technical report, {OpenAI}.

\bibitem[{Regier and Gahl(2004)}]{regier2004}
Terry Regier and Susanne Gahl. 2004.
\newblock \href {https://doi.org/10.1016/j.cognition.2003.12.003} {Learning the unlearnable: The role of missing evidence}.
\newblock \emph{Cognition}, 93(2):147--155.

\bibitem[{Roland et~al.(2007)Roland, Dick, and Elman}]{roland2007}
Douglas Roland, Frederic Dick, and Jeffrey~L. Elman. 2007.
\newblock \href {https://doi.org/10.1016/j.jml.2007.03.002} {Frequency of basic {{English}} grammatical structures: {{A}} corpus analysis}.
\newblock \emph{Journal of Memory and Language}, 57(3):348--379.

\bibitem[{Theakston(2004)}]{theakston2004}
Anna~L Theakston. 2004.
\newblock \href {https://doi.org/10.1016/j.cogdev.2003.08.001} {The role of entrenchment in children's and adults' performance on grammaticality judgment tasks}.
\newblock \emph{Cognitive Development}, 19(1):15--34.

\bibitem[{Vaswani et~al.(2017)Vaswani, Shazeer, Parmar, Uszkoreit, Jones, Gomez, Kaiser, and Polosukhin}]{vaswani2017}
Ashish Vaswani, Noam Shazeer, Niki Parmar, Jakob Uszkoreit, Llion Jones, Aidan~N Gomez, {\L}ukasz Kaiser, and Illia Polosukhin. 2017.
\newblock Attention is {{All}} you {{Need}}.
\newblock In \emph{Advances in {{Neural Information Processing Systems}} ({{NIPS}} 2017)}, volume~30. {Curran Associates, Inc.}

\bibitem[{Warstadt et~al.(2020)Warstadt, Parrish, Liu, Mohananey, Peng, Wang, and Bowman}]{warstadt2020}
Alex Warstadt, Alicia Parrish, Haokun Liu, Anhad Mohananey, Wei Peng, Sheng-Fu Wang, and Samuel~R. Bowman. 2020.
\newblock \href {https://doi.org/10.1162/tacl_a_00321} {{{BLiMP}}: {{The Benchmark}} of {{Linguistic Minimal Pairs}} for {{English}}}.
\newblock \emph{Transactions of the Association for Computational Linguistics}, 8:377--392.

\bibitem[{Warstadt et~al.(2019)Warstadt, Singh, and Bowman}]{warstadt-etal-2019-neural}
Alex Warstadt, Amanpreet Singh, and Samuel~R. Bowman. 2019.
\newblock \href {https://doi.org/10.1162/tacl_a_00290} {Neural network acceptability judgments}.
\newblock \emph{Transactions of the Association for Computational Linguistics}, 7:625--641.

\bibitem[{Wei et~al.(2021)Wei, Garrette, Linzen, and Pavlick}]{wei-etal-2021-frequency}
Jason Wei, Dan Garrette, Tal Linzen, and Ellie Pavlick. 2021.
\newblock \href {https://doi.org/10.18653/v1/2021.emnlp-main.72} {Frequency effects on syntactic rule learning in transformers}.
\newblock In \emph{Proceedings of the 2021 Conference on Empirical Methods in Natural Language Processing}, pages 932--948, Online and Punta Cana, Dominican Republic. Association for Computational Linguistics.

\bibitem[{Zaenen(1993)}]{zaenen1993}
Annie Zaenen. 1993.
\newblock \href {https://doi.org/10.1007/978-94-011-1972-6_9} {Unaccusativity in {{Dutch}}: {{Integrating Syntax}} and {{Lexical Semantics}}}.
\newblock In James Pustejovsky, editor, \emph{Semantics and the {{Lexicon}}}, Studies in {{Linguistics}} and {{Philosophy}}, pages 129--161. {Springer Netherlands}, {Dordrecht}.

\bibitem[{Zwicky(1987)}]{zwicky1987}
Arnold~M. Zwicky. 1987.
\newblock \href {https://doi.org/10.1515/ling.1987.25.4.639} {Slashes in the passive}.
\newblock \emph{Linguistics}, 25(4).

\end{thebibliography}

\appendix
\section{Stimuli}
\label{sec:stimuli}
\subsection{Test sentence frames}
\label{sec:test-sentences}

\begin{tabular}{ll}
     \toprule
     Verb class & Sentence frame\\
     \midrule
     \multirow{5}{*}{Advantage}& Your investment \gap{} the community. \\
     & The exercise \gap{} his fitness.\\
     & Our friendship \gap{} my life. \\
     & The law \gap{} these workers. \\
     & The treaty \gap{} both countries. \\
     \midrule
     \multirow{5}{*}{Price} & Your dish \gap{} ninety dollars.\\
     & The painting \gap{} a fortune. \\
     & The tickets \gap{} a lot of money.\\
     & Your book \gap{} thirty dollars. \\
     & His actions \gap{} the medal. \\
     \midrule
     \multirow{5}{*}{Ooze} & My friend \gap{} confidence.\\
     & The lightbulb \gap{} some light.\\
     & That machine \gap{} a sound. \\
     & The teacher \gap{} wisdom.\\
     & The trash \gap{} an odor.\\
     \midrule 
     \multirow{5}{*}{Estimation} & Your drawing \gap{} her likeness.\\
     & Your friend \gap{} my brother.\\
     & The character \gap{} the author.\\
     & Her son \gap{} her father.\\
     & The copy \gap{} the original.\\
     \midrule
     \multirow{5}{*}{Duration} & The journey \gap{} three days.\\
     & My meeting \gap{} two hours.\\
     & The interview \gap{} some time. \\
     & Her speech \gap{} seventeen minutes. \\
     & His trek \gap{} a month.\\
     \bottomrule
\end{tabular}
\vfill{}
\pagebreak

\subsection{Agent-patient sentences}
\label{sec:control-sentences}
\begin{tabular}{ll}
\toprule
Verb & Active sentence\\
\midrule
\multirow{5}{*}{hit} & My brother hit your friend.\\
& Your sister hit the target.\\
& The child hit the ball.\\
& A boy hit my bag.\\
& A monkey hit the toy.\\
\midrule
\multirow{5}{*}{pushed} & My brother pushed a child.\\
& The mother pushed my toy.\\
& A boy pushed the cup.\\
& A child pushed the bag.\\
& Your sister pushed your friend.\\
\midrule
\multirow{5}{*}{washed} & A boy washed the cup.\\
& A child washed the bag.\\
& My sister washed a towel.\\
& My brother washed my plate.\\
& Your mother washed my toy.\\
\midrule
\multirow{5}{*}{dropped} & My brother dropped my plate.\\
& The mother dropped my toy.\\
& A boy dropped the cup.\\
& A child dropped the bag.\\
& Your sister dropped a book.\\
\midrule
\multirow{5}{*}{carried} & A boy carried my bag.\\
& Your mother carried the child.\\
& My brother carried your friend.\\
& The dog carried the toy.\\
& The donkey carried the load.\\
\bottomrule
\end{tabular}

\subsection{Experiencer-theme sentences}
\begin{tabular}{ll}
\toprule
     Verb & Active sentence\\
     \midrule
\multirow{5}{*}{saw} & My brother saw your friend.\\
& Your dog saw the toy.\\
& Your sister saw a book.\\
& A boy saw my bag.\\
& The child saw a monkey.\\
\midrule
\multirow{5}{*}{heard} & A boy heard the sound.\\
& The child heard the rules.\\
& My brother heard your friend.\\
& Your dog heard the toy.\\
& Your sister heard a squeak.\\
\midrule
\multirow{5}{*}{knew} & My brother knew your friend.\\
& Your dog knew my cat.\\
& Your sister knew my brother.\\
& A boy knew my mother.\\
& The mother knew the dog.\\
\midrule
\multirow{5}{*}{liked} & A boy liked the game.\\
& The child liked a monkey.\\
& My brother liked your friend.\\
& Your dog liked the toy.\\
& Your sister liked a book.\\
\midrule
\multirow{5}{*}{remembered} & My brother remembered your friend.\\
& Your dog remembered my toy.\\
& Your sister remembered a book.\\
& A boy remembered the game.\\
& The child remembered the rules.\\
\bottomrule
\end{tabular}
\end{document}